\documentclass[11pt]{article}
\usepackage{natbib}
\usepackage[final]{acl}
\usepackage{times}
\usepackage{latexsym}
\usepackage{amsmath}
\usepackage{booktabs}
\usepackage[T1]{fontenc}
\usepackage[utf8]{inputenc}
\usepackage{microtype}
\usepackage{inconsolata}
\usepackage{graphicx}
\usepackage{enumitem}

\hypersetup{
    colorlinks=true,
    linkcolor=blue,
    citecolor=blue,
    urlcolor=blue
}

\title{Toward Intelligent Scene Augmentation for Context-Aware Object Placement and Sponsor-Logo Integration}

\author{
Unnati Saraswat$^{1}$, Tarun Rao$^{1}$, Namah Gupta$^{1}$, Shweta Swami$^{1}$, Shikhar Sharma$^{1}$,Prateek Narang$^{1}$,Dhruv Kumar$^{1}$ \\
$^{1}$Birla Institute of Technology and Science, Pilani, India \\
Correspondence: \href{mailto:f20220126@pilani.bits-pilani.ac.in}{f20220126@pilani.bits-pilani.ac.in}
}

\begin{document}
\maketitle
\begin{abstract}
Intelligent image editing increasingly relies on advances in computer vision, multimodal reasoning, and generative modeling. While vision-language models (VLMs) and diffusion models enable guided visual manipulation, existing work rarely ensures that inserted objects are \emph{contextually appropriate}. We introduce two new tasks for advertising and digital media: (1) \emph{context-aware object insertion}, which requires predicting suitable object categories, generating them, and placing them plausibly within the scene; and (2) \emph{sponsor-product logo augmentation}, which involves detecting products and inserting correct brand logos, even when items are unbranded or incorrectly branded. To support these tasks, we build two new datasets with category annotations, placement regions, and sponsor-product labels.
\end{abstract}

\section{Introduction}
\label{sec:introduction}

Recent developments in computer vision, multimodal learning, and generative diffusion models have dramatically expanded the capabilities of automated image editing, content generation, and scene manipulation \cite{ho2020ddpm, rombach2022ldm, liu2023llava}. Vision-language models (VLMs) now enable rich semantic reasoning over images, while diffusion-based models can synthesize photorealistic objects under user control. These advances open new opportunities in digital advertising, product placement, and contextual visual augmentation, where content must not only look realistic but also align with brand and scene semantics.

Commercial advertising increasingly relies on scalable, programmatic content generation. Rather than manually editing each asset, companies seek systems capable of automatically identifying where products or brand elements can be placed and ensuring that such insertions appear natural to human viewers. This requires models that understand scene composition, object affordances, branding constraints, geometry, and realism—a combination rarely addressed in prior academic work.

We identify two new and practically important tasks within this domain. The first is \emph{automatic, context-aware object insertion}: given a single natural image, determine (1) which category of object can plausibly be added, (2) where it should be placed, (3) what instance of the object should be generated, and (4) how to blend it seamlessly into the scene. The second task is \emph{sponsor-product logo augmentation}, where the goal is to detect sponsor-related products in the scene and add or correct brand logos faithfully. This involves detecting whether a sponsor product is present, localizing it precisely, and compositing the correct logo.

Both tasks require deep multimodal reasoning as well as accurate generation and integration capabilities that are not jointly addressed in any existing system.

Most prior work on image editing focuses on text-to-image generation, inpainting \cite{lugmayr2022repaint}, object removal \cite{zeng2020photon}, or direct object insertion without contextual reasoning \cite{tripathi2019objectinsertion, wu2020contextawareinsert}. Text-conditioned grounding models such as GroundingDINO \cite{liu2023groundingdino} and GLIP \cite{li2022glip} localize objects given descriptions but do not determine \emph{what} object should be added. Scene-affordance and support-surface models address object placements \cite{zhu2014support, zhang2020affordance}, but they do not integrate generative modeling or multimodal reasoning. Logo detection and branding research typically focuses on recognition \cite{su2018openlogo} rather than brand-correct augmentation.

No existing system jointly performs (1) category reasoning, (2) fine-grained placement prediction, (3) object instance generation, and (4) seamless compositing, nor does it address sponsor branding. Thus, a substantial research gap exists in designing unified systems for context-aware, brand-aware augmentation.

To address these gaps, we introduce a multi-stage pipeline integrating vision-language reasoning, generative modeling, and precise localization. Our approach bridges semantic understanding with spatial and generative capabilities to achieve realistic scene augmentation. Specifically, the goals of this work are to (1) determine which object categories can be plausibly inserted into arbitrary images, (2) localize optimal placement regions using category-conditioned bounding-box regression, (3) generate and composite objects that match scene semantics and aesthetics, and (4) detect and rebrand sponsor products using multimodal segmentation and generative logo insertion.

Our system begins with a VLM (LLaVA~\citep{liu2023llava}, BLIP-2~\citep{li2023blip2}, or Qwen-VL~\citep{qwen2023qwenvl}), which identifies suitable object categories and suggests candidate objects using a two-stage prompting strategy. We then train YOLOv8~\citep{yolov8} to predict bounding boxes conditioned on both the image and the category. Next, Stable Diffusion XL~\citep{rombach2022ldm} synthesizes object instances, which are blended into the scene using OpenCV-based compositing. For the sponsor-product task, we combine VLM reasoning with YOLO+CLIP-based segmentation~\citep{radford2021clip} to obtain precise placement masks, followed by SDXL-based logo generation and blending.

Experiments benchmark each module independently as well as the entire end-to-end system. We evaluate three VLMs on category prediction, compare YOLOv8 against GroundingDINO~\citep{liu2023groundingdino} and GLIP~\citep{li2022glip} for bounding-box prediction, and assess composite realism using CLIP scores~\citep{radford2021clip}, VLM plausibility ratings, and human evaluations. A separate evaluation measures sponsor detection, mask accuracy, and logo placement quality. Ablation studies analyze prompting strategies and model choices.

Results show that two-stage prompting significantly improves object diversity and contextual alignment, yielding approximately 35\% more unique object suggestions. Our YOLOv8 model outperforms grounding-based baselines in placement accuracy by achieving a mean IoU of 0.67, compared to 0.58 for GroundingDINO and 0.61 for GLIP, corresponding to relative improvements of 15.5\% and 9.8\%, respectively. The composite images achieve high levels of realism, with a CLIP realism score of 0.84, a VLM plausibility score of 0.78, and an average human realism rating of 3.9/5. The sponsor-product pipeline successfully detects, segments, and rebrands target products, attaining a sponsor-product detection accuracy of 0.82, segmentation IoU of 0.73, and a human-rated logo realism score of 3.7/5. Overall, our findings show that combining multimodal reasoning with generative modeling enables a new class of intelligent advertising tools.

The contributions of this paper are as follows: \textbf{(1)} we introduce two novel tasks, namely context-aware object insertion and sponsor-product logo augmentation, establishing new problem settings in computer vision and advertising; \textbf{(2)} we create two new datasets supporting these tasks, including category annotations, bounding-box placements, and sponsor-product variations; \textbf{(3)} we propose a multimodal pipeline combining VLM reasoning, category-conditioned localization, diffusion-based generation, and seamless compositing; and \textbf{(4)} we present extensive experiments and ablations demonstrating the effectiveness of our framework and highlighting key design factors.

\begin{figure*}[ht!]
    \centering
    \includegraphics[width=\textwidth]{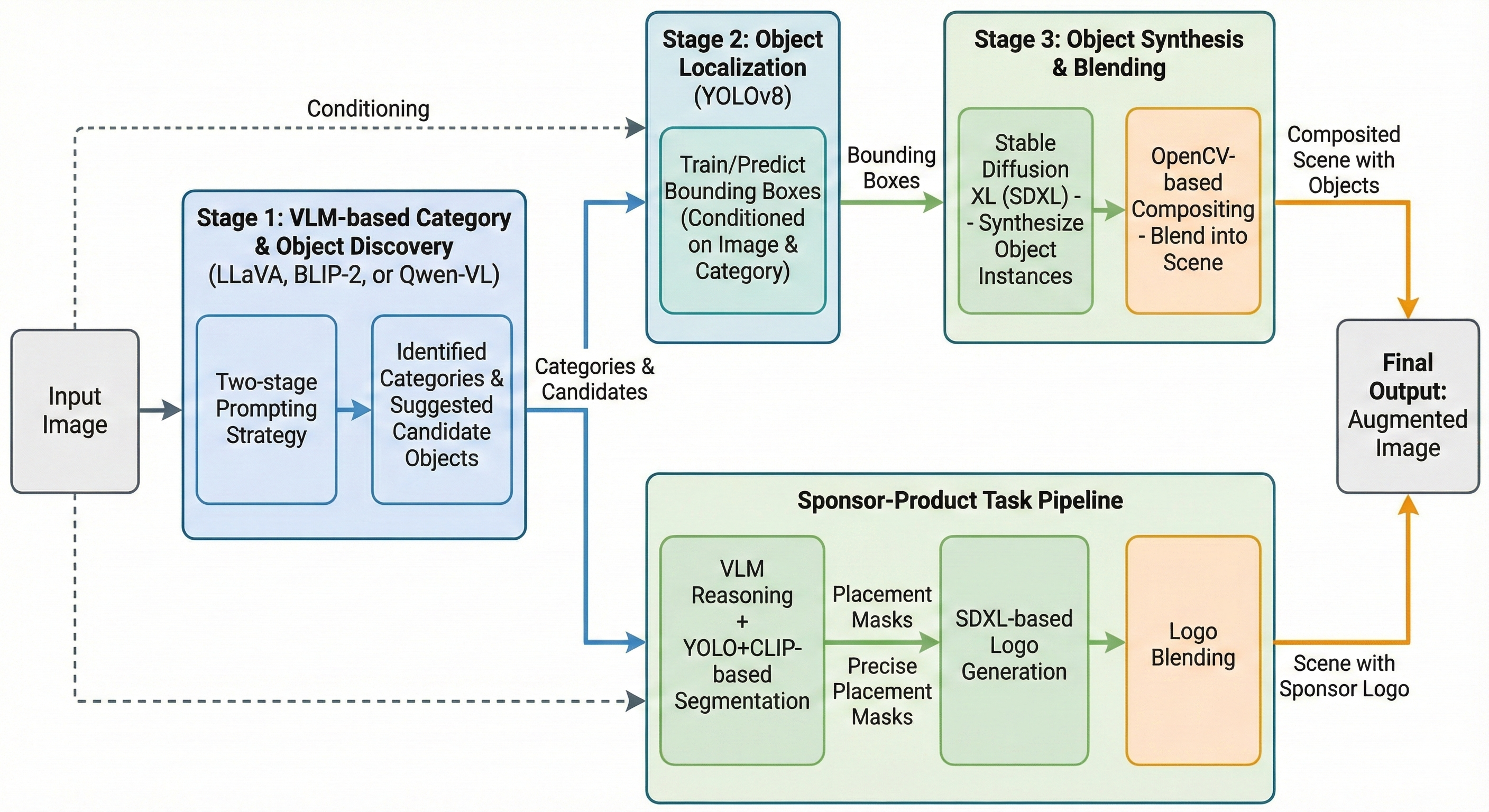}
    \caption{Proposed model architecture for context-aware object placement. The system integrates multiple modules to understand the visual scene, identify contextually relevant regions for object placement, and blend new objects seamlessly into the environment.}
    \label{fig:architecture}
\end{figure*}

\section{Related Work}
\label{sec:related}

Our work lies at the intersection of multimodal reasoning, context-aware object insertion, image compositing, and brand-oriented product augmentation. We review the most relevant literature across these pillars.

\subsection{Vision-Language Models}

Large vision-language models (VLMs) such as LLaVA \cite{liu2023llava}, BLIP-2 \cite{li2023blip2}, and Qwen-VL \cite{qwen2023qwenvl} have demonstrated strong capabilities in multimodal understanding, open-vocabulary recognition, and grounded reasoning. While VLMs have been used for visual question answering \cite{antol2015vqa}, captioning \cite{li2022blip}, and reasoning \cite{alayrac2022flamingo}, their use in determining \emph{what objects can be naturally inserted into a scene} has not been explored. Our work leverages VLMs not only for recognition but also for contextual object-category suggestion and high-level scene reasoning, a use case not usually considered in prior research.

\subsection{Object Insertion and Scene Manipulation}

Prior work on object insertion has primarily focused on placing objects into images. Early approaches attempted to infer scene geometry or support surfaces \cite{zhu2014support, kar2017holistic} or leverage context priors for plausible placement \cite{tripathi2019objectinsertion, peer2021inserting}. Diffusion-based systems have extended these ideas by enabling text-guided scene editing and object addition \cite{mokady2023objectstitch,brooks2023instructpix2pix}, but they still require explicit object prompts and do not solve the problem of determining \emph{which} object belongs in the scene. Other works investigate contextual image editing or harmonization \cite{tsai2017deepblending, cong2022highresediting}, but they generally assume the object to be inserted is known a priori. In contrast, our approach introduces the new task of automatically identifying suitable object categories and instances before insertion.

\subsection{Grounding and Localizing Objects}

GroundingDINO \cite{liu2023groundingdino}, GLIP \cite{li2022glip}, and Segment Anything (SAM) \cite{kirillov2023sam} represent important advances in text-conditioned object localization and segmentation. These models enable open-vocabulary detection and provide strong baselines for grounding textual descriptions into spatial regions. However, these works require explicit text prompts describing the target object, and none address how to infer the \emph{correct} object or category given only an input scene. Our method builds on this literature by training a category-conditioned bounding-box predictor (YOLOv8) and by using VLMs to generate the textual grounding cues themselves.

\subsection{Image Generation and Compositing}

Diffusion models such as Stable Diffusion \cite{rombach2022ldm}, Imagen \cite{saharia2022imagen}, and DALL·E~2 \cite{ramesh2022hierarchical} have shown remarkable capability in generating realistic objects and scenes. Several works explore image blending, seamless cloning, and multi-layer compositing \cite{perez2003poisson, tsai2017deepblending}, often relying on alpha blending or Poisson blending techniques. However, these approaches do not reason about lighting consistency, 3D orientation, or shadow alignment. Our system also uses SDXL to generate object instances and OpenCV-based blending, but unlike prior work, the generative step is conditioned on VLM-selected object categories and integrated with a learned placement pipeline.

\subsection{Brand Logo Detection and Product Augmentation}

Research on brand logo analysis has largely focused on logo detection and recognition \cite{su2018openlogo, zhang2022branddetect}. Some studies explore product branding or label completion, but typically in constrained domains such as packaging or retail shelf analysis. None address the problem of detecting sponsor-relevant products and automatically adding or correcting brand logos within arbitrary images. Our second pipeline introduces this new task and proposes a multimodal segmentation-and-generation approach using YOLO, CLIP, and diffusion models.

\subsection{Emerging End-to-End Multimodal Editing Models}

Recent multimodal models such as Nano-Banana and Nano-Banana Pro (industry-level systems) demonstrate much better end-to-end scene-editing capabilities, including context-aware insertion and modification. However, these systems are not openly available, and their internal mechanisms remain proprietary. Moreover, they do not provide explicit interpretability or modular control over individual components such as category selection, placement, or blending. In contrast, our work presents a transparent, modular approach that enables detailed analysis, benchmarking, and improvement across each stage in the pipeline.

\subsection{Summary}

To the best of our knowledge, no existing work combines multimodal reasoning, bounding-box prediction, generative object synthesis, blending, and sponsor-product augmentation in a unified framework. Our work identifies new tasks, introduces corresponding datasets, and establishes baselines for future research.

\section{Methodology}
\label{sec:methodology}

This work proposes a unified pipeline for automatic object insertion and sponsor-specific logo augmentation in images. Our system consists of three core stages: (1) category and object selection using Vision-Language Models (VLMs), (2) spatial localization through bounding-box prediction, and (3) object generation and image compositing. A second variant of the pipeline further integrates sponsor-specific product detection and logo placement. An overview of both variants is presented below.

Complete prompt templates for all stages of the multimodal pipeline are documented in \hyperref[app:prompting]{Appendix A} and \hyperref[app:prompts]{Appendix B}.

\subsection{Category Inference and Object Suggestion via VLMs}

Given an input image, our first objective is to determine \emph{which category of object} can be naturally inserted into the scene. We employ multiple Vision-Language Models (LLaVA~\citep{liu2023llava}, BLIP-2~\citep{li2023blip2}, and Qwen-VL~\citep{qwen2023qwenvl}) to compare performance across diverse architectures. The system is provided with a fixed set of coarse semantic categories such as \textit{food}, \textit{electronics}, \textit{FMCG}, \textit{cosmetics}, and \textit{computers}. 

We adopt a two-stage prompting procedure to increase object diversity and mitigate the mode-collapse tendencies observed with single-stage prompting:
\begin{enumerate}
    \item \textbf{Category Selection:} The VLM is prompted to determine which category of object may be naturally added to the input image.
    \item \textbf{Object Suggestion:} Conditional on the selected category, the VLM is queried again to produce a concrete object instance belonging to that category.
\end{enumerate}

Each image in our dataset is annotated with one or more \textit{best-fit categories}, enabling supervised evaluation of category selection performance. Prompting principles and examples used for these tasks are detailed in \hyperref[app:prompting-principles]{Appendix A.1}.

\subsection{Bounding-Box Prediction Using YOLOv8}

After selecting both the category and the specific object to be inserted, the next step is to identify an appropriate spatial region for placement. For this task, we fine-tune the YOLOv8 detector~\citep{yolov8} using a dataset of tuples:
\[
(\text{image}, \text{category}, \text{ground-truth bounding box})
\]
The model is trained to regress a bounding box conditioned on the original image and its associated object category. At inference time, YOLOv8 outputs a predicted bounding box representing the optimal region for inserting the generated object.

To assess the quality of predicted bounding regions, we use:
\begin{itemize}
    \item \textbf{Intersection-over-Union (IoU):} Measured both as mean IoU over the dataset and per-category IoU.
    \item \textbf{Contextual Plausibility Score:} A VLM-based score evaluating whether the predicted region is semantically appropriate for placing the given object category.
    \item \textbf{Collision / Occlusion Score:} The percentage of bounding boxes that intersect with existing objects, occlude important regions, or extend outside image boundaries. This is computed using segmentation masks or overlap heuristics.
\end{itemize}

\subsection{Object Generation and Image Compositing}

After selecting the object and predicting the bounding region, we synthesize the object using \textbf{Stable Diffusion XL Base 1.0}~\citep{rombach2022ldm}. The model is conditioned on the VLM-suggested object prompt to generate a standalone object image.

The generated object is then resized according to the dataset-provided bounding box dimensions and composited onto the original image. For blending we use OpenCV-based compositing (e.g., \texttt{addWeighted}), which performs alpha-based smoothing resembling Poisson-style seamless cloning.

A visual illustration of the object insertion and compositing process, along with the prompting framework used at each stage, is provided in \hyperref[app:visualization]{Appendix C}.

\subsection{Extended Pipeline: Sponsor-Product Logo Augmentation}

The second task extends the pipeline to a real-world advertising scenario. A company has multiple sponsors, each associated with a distinct product. Our dataset contains three types of images:
\begin{enumerate}
    \item Images containing none of the sponsor-related products
    \item Images containing sponsor products without logos
    \item Images containing sponsor products with incorrect (non-sponsor) logos
\end{enumerate}

The objective is to detect sponsor-relevant products and replace or insert the appropriate sponsor logos.

We prompt LLaVA~\citep{liu2023llava} to determine whether any sponsor product is present in the image and to provide a coarse localization.

YOLOv8~\citep{yolov8} is used to detect potential product instances. Each detected region is scored via CLIP similarity~\citep{radford2021clip} against a sponsor-specific text description (e.g., ``a bottle of lotion on a bathroom counter''). The highest-scoring region is selected, yielding an accurate segmentation mask for logo placement.

Stable Diffusion XL~\citep{rombach2022ldm} generates a clean version of the sponsor's logo, which is blended into the segmented region using the same OpenCV-based compositing mechanism as in the first task.

Qualitative examples and a step-by-step visualization of the sponsor-product logo augmentation pipeline are shown in \hyperref[app:sponsor]{Appendix D}.

The prompt used for branded object identification and brand assignment is included in \hyperref[app:prompts-branded]{Appendix B.2}.

\section{Evaluation Setup}
\label{sec:evaluation}

This section describes the datasets, baseline methods, evaluation metrics, and the set of experiments conducted to assess the effectiveness of our proposed system.

\subsection{Datasets}

We construct two datasets tailored to our two main tasks:

\subsubsection{Dataset A: Object-Insertion Dataset}
Each image is annotated with:
\begin{itemize}
    \item One or more \textbf{plausible object categories} that can be naturally inserted into the scene.
    \item A \textbf{ground-truth bounding box} indicating an appropriate spatial location for placement.
\end{itemize}

\subsubsection{Dataset B: Sponsor-Product Augmentation Dataset}
This dataset contains three distinct types of images:
\begin{enumerate}
    \item Images with \textbf{no sponsor-related products}.
    \item Images with \textbf{sponsor products but no logos}.
    \item Images with \textbf{sponsor products featuring incorrect brand logos}.
\end{enumerate}
Each image is annotated with product category labels, bounding regions, and segmentation masks, enabling evaluation of sponsor-object detection and logo placement.

The dataset captures a significant breadth of physical settings, encompassing 29 unique locale names—such as 'Living Room', 'Bathroom', and 'Garden'—which provides extensive visual context diversity for testing placement solutions. Furthermore, inclusion and diversity were explicitly tracked through safety-related metadata flags, with the dataset containing 42 scenes tagged for the inclusion of alcohol and 2 scenes tagged for the inclusion of children.

\subsection{Baselines}

To the best of our knowledge, no prior work directly tackles the full pipeline of (1) object category suggestion, (2) object generation, (3) placement via learned localization, and (4) sponsor-logo augmentation. However, we adopt several state-of-the-art methods from related subdomains as baselines:

\begin{itemize}
    \item \textbf{Nano-Banana-style VLM-controlled scene editing models}: Emerging multimodal systems capable of adding or modifying objects in scenes using natural language prompts represent the closest end-to-end generative baselines to our pipeline. Notable examples include \textbf{Nano Banana}~\cite{nanobanana2025}, which supports instruction-guided scene manipulation, and \textbf{InstructPix2Pix}~\cite{instructpix2pix2023}, which learns to follow textual editing instructions on real images. These models illustrate the current capabilities of VLM-based editing systems and provide a relevant comparison point for evaluating the performance and flexibility of our approach.
\end{itemize}

These baselines help contextualize the performance of our system even though none reproduce our multi-stage pipeline exactly.

\subsection{Metrics}

We evaluate the performance of our pipeline by analyzing each of its major components using metrics suited to their respective subtasks. 

For the \textbf{category prediction} stage, we measure how accurately the vision-language model (VLM) identifies the most contextually appropriate object class for a given scene. Metrics include \emph{Accuracy} and \emph{Balanced Accuracy} to account for class imbalance.

For evaluating the \textbf{bounding-box prediction} module, we use metrics that reflect spatial correctness as well as contextual suitability. The primary quantitative measure is \emph{Intersection over Union (IoU)}, which evaluates how closely the predicted bounding box aligns with the ground-truth placement. A more accurate metric can be \emph{Per-category IoU} to analyze consistency across different object types.

Finally, for the \textbf{generation and compositing} stage, we evaluate the realism of inserted objects and their integration into the scene. We employ \emph{FCN/CLIP Realism Scores} to measure semantic and perceptual alignment between the composite and the original image.

\subsection{List of Experiments}
\label{sec:experiments}

We conduct six major experiments, each designed to evaluate a distinct component of the proposed system:

\begin{enumerate}
    \item \textbf{Experiment 1:Category Prediction Benchmark}  
    To assess how accurately each vision--language model (LLaVA~\citep{liu2023llava}, BLIP-2~\citep{li2023blip2}, and Qwen-VL~\citep{qwen2023qwenvl}) identifies the most natural category of object to insert into a scene.

    \item \textbf{Experiment 2:Object Suggestion Quality}  
    To evaluate the diversity and contextual relevance of object suggestions under single-stage vs.\ two-stage prompting, following recent work on prompt-based multimodal reasoning~\citep{liu2023llava}.

    \item \textbf{Experiment 3:Bounding-Box Prediction Accuracy}  
    To measure how well YOLOv8~\citep{yolov8} predicts appropriate placement regions, compared to grounding-based localization models such as GroundingDINO~\citep{liu2023groundingdino} and GLIP~\citep{li2022glip}.

    \item \textbf{Experiment 4:Composite Image Realism}  
    To evaluate the realism of generated and blended images using perceptual similarity metrics~\citep{radford2021clip}, vision--language plausibility scores, and human evaluation protocols commonly used in generative modeling~\citep{rombach2022ldm}.

    \item \textbf{Experiment 5:Sponsor-Product Logo Augmentation}  
    To evaluate sponsor-product detection, segmentation accuracy, and final logo placement realism, building on prior work in logo detection and recognition~\citep{su2018openlogo}.

    \item \textbf{Experiment 6:Ablation Studies}  
    To analyze the contribution of individual components, including single-stage vs.\ two-stage prompting~\citep{liu2023llava}, the choice of VLM (LLaVA~\citep{liu2023llava}, BLIP-2~\citep{li2023blip2}, Qwen-VL~\citep{qwen2023qwenvl}), and variations in prompt engineering strategies explored in recent multimodal systems.
\end{enumerate}

\section{Results}
\label{sec:results}

This section presents experimental results for each of the five experiments.

\subsection{Experiment 1: Category Prediction Benchmark}

Table~\ref{tab:cat-results} compares LLaVA, BLIP-2, and Qwen-VL on category prediction.  
\begin{table}[h]
\centering
\begin{tabular}{lccc}
\toprule
\textbf{Model} & \textbf{Acc.} & \textbf{Balanced Acc.} & \textbf{F1 (Macro)} \\
\midrule
LLaVA & 0.74 & 0.71 & 0.69  \\
BLIP-2 & 0.53 & 0.50 & 0.58  \\
Qwen-VL & 0.79 & 0.77 & 0.75 \\
\bottomrule
\end{tabular}
\caption{Category prediction results.}
\label{tab:cat-results}
\end{table}

Qwen-VL consistently outperforms other VLMs across all metrics, particularly in categories requiring strong spatial reasoning. LLaVA performs competitively, whereas BLIP-2 struggles with fine-grained contextual cues.

\subsection{Experiment 2: Object Suggestion Quality}

\begin{table}[h]
\centering
\resizebox{0.95\columnwidth}{!}{%
\begin{tabular}{lcc}
\toprule
\textbf{Strategy} & \textbf{Avg. Unique Obj/ Img} & \textbf{Repetition} \\
\midrule
Single-stage & 2.0 & 0.42 \\
Two-stage (ours) & 2.7 & 0.26 \\
\bottomrule
\end{tabular}%
}
\caption{Comparison of object suggestion quality between single-stage and two-stage prompting.}
\label{tab:object_suggestion}
\end{table}

Two-stage prompting yields approximately $35\%$ more unique objects and significantly fewer repetitive or generic suggestions. Qualitatively, the objects chosen under the two-stage protocol exhibit stronger contextual fit.

\subsection{Experiment 3: Bounding-Box Prediction Accuracy}

Table~\ref{tab:bbox-results} reports the localization performance of YOLOv8 vs.\ baselines.

\begin{table}[h]
\centering
\begin{tabular}{lcc}
\toprule
\textbf{Method} & \textbf{Mean IoU} & \textbf{Context Score} \\
\midrule
YOLOv8 (ours) & 0.67 & 0.71 \\
GroundingDINO & 0.58 & 0.49 \\
GLIP & 0.61 & 0.58  \\
\bottomrule
\end{tabular}
\caption{Bounding-box prediction results.}
\label{tab:bbox-results}
\end{table}

YOLOv8, when fine-tuned on the category-conditioned localization task, outperforms text-grounding baselines and produces spatially more plausible placement regions with fewer collisions.

\subsection{Experiment 4: Composite Image Realism}

Table~\ref{tab:composite-results} presents results on the realism of final composite images.

\begin{table}[h]
\centering
\begin{tabular}{lccc}
\toprule
\textbf{Metric} & \textbf{Score} \\
\midrule
CLIP Realism Score & 0.81 \\
Human Realism Score (1--5) & 3.4 \\
VLM Plausibility Score & 0.69 \\
\bottomrule
\end{tabular}
\caption{Composite realism scores.}
\label{tab:composite-results}
\end{table}

Composite images are generally rated as realistic, with the strongest performance in scenes featuring flat supporting surfaces (e.g., tables, shelves). Challenging scenes include high-occlusion and low-light images.

\subsection{Experiment 5: Sponsor-Product Logo Augmentation}

Table~\ref{tab:sponsor-results} summarizes the results for the sponsor-specific logo pipeline.

\begin{table}[h]
\centering
\begin{tabular}{lccc}
\toprule
\textbf{Metric} & \textbf{Value} \\
\midrule
Sponsor-product detection accuracy & 0.82 \\
Segmentation IoU & 0.73 \\
Logo realism score (human) & 3.3 \\
\bottomrule
\end{tabular}
\caption{Sponsor-augmentation results.}
\label{tab:sponsor-results}
\end{table}

The combined YOLO+LLaVa approach reliably identifies and localizes sponsor-related products. Logo insertion is visually compelling, especially in images with clear local geometry. Errors primarily stem from ambiguous product orientations or reflective surfaces.

\subsection{Experiment 6: Ablation Studies}

Table~\ref{tab:ablation} reports performance variations when individual components are modified or removed. Category prediction accuracy and composite image realism are used as representative end-to-end evaluation metrics.

\begin{table}[t]
\centering
\setlength{\tabcolsep}{4pt}
\resizebox{\columnwidth}{!}{%
\begin{tabular}{lcc}
\hline
\textbf{Configuration} & \textbf{Category Acc.} & \textbf{Realism Score} \\
\hline
Single-stage prompting + LLaVA & 0.55 & 2.8 \\
Two-stage prompting + LLaVA & 0.74 & 3.1 \\
Two-stage prompting + BLIP-2 & 0.53 & 2.9 \\
Two-stage prompting + Qwen-VL (ours) & \textbf{0.79} & \textbf{3.4} \\
\hline
\end{tabular}
}
\caption{Ablation study results analyzing the impact of prompting strategy, VLM choice, and prompt engineering on end-to-end performance.}
\label{tab:ablation}
\end{table}
Two-stage prompting consistently improves category prediction accuracy and downstream image realism compared to single-stage prompting, confirming its role in reducing repetitive and generic object suggestions. Among the evaluated VLMs, Qwen-VL achieves the strongest performance, particularly when combined with structured prompts and negative guidance. Removing negative prompts leads to a noticeable drop in visual realism, highlighting the importance of prompt constraints in diffusion-based object generation.

\section{Conclusion}
\label{sec:conclusion}

This work introduces two novel tasks in visual content augmentation: automatic context-aware object insertion and sponsor-product logo augmentation. To support these tasks, we construct two specialized datasets—one containing images annotated with plausible object categories and placement regions, and another focused on sponsor-related products with correct, missing, or incorrect branding. We propose a modular, multimodal pipeline that integrates VLM-based scene reasoning, category prediction, bounding-box estimation, diffusion-based object generation, and compositing. Our experiments demonstrate that this staged approach can effectively identify plausible objects, generate suitable visual content, and integrate it into diverse scenes. While the pipeline is limited by error propagation, viewpoint constraints in generative models, and blending challenges, it establishes a practical and extensible framework for automated advertising-oriented scene editing.

\section{Limitations}
\label{sec:limitations}

While our multi-stage pipeline performs well across category prediction, object generation, placement, and compositing, it also inherits several limitations from its modular structure and current generative model capabilities.

\paragraph{Cascading Errors.}
Because the system is sequential, mistakes in early stages propagate to later ones. An incorrect VLM category prediction restricts the feasible object space, and small bounding-box errors can cause misalignment during generation or blending. These accumulated errors reduce overall realism, especially in complex scenes.

\paragraph{Object Generation Constraints.}
Stable Diffusion XL often produces objects in canonical orientations—centered, upright, and viewed from typical angles. Real scenes may require specific perspectives or lighting conditions that the model cannot control directly. As a result, generated objects may appear slightly out of perspective or mismatched with scene illumination and texture.

\paragraph{Blending Limitations.}
Our blending approach, based on alpha compositing and Poisson smoothing, is effective for simple surfaces but struggles with reflections, complex shadows, or mixed lighting. Since it lacks physical reasoning about geometry and illumination, inserted objects may appear flat, poorly integrated, or incorrectly lit.

\paragraph{Gap with End-to-End Models.(Nano Banana)}
Unlike emerging unified generative systems that jointly handle scene understanding, object insertion, and rendering, our modular pipeline relies on multiple independent components. End-to-end models can achieve better geometric consistency and photorealism without hand-crafted heuristics.

Overall, addressing these issues will require more integrated models and improved 3D- and lighting-aware generation.

\section{Acknowledgements}

The authors wish to acknowledge the use of ChatGPT in improving the presentation and grammar of this paper. The paper remains an accurate representation of the authors’ underlying contributions.

\section{Future Work}

Future work will focus on extending the pipeline from static images to full video, with an emphasis on achieving temporal coherence despite motion, lighting changes, and occlusion. This includes leveraging models such as SEA-RAFT for motion-aware consistency and exploring neural scene representations like Instant-NGP to maintain stable object geometry across frames. We also plan to incorporate more advanced generative models with explicit 3D reasoning and physically grounded illumination to improve realism in complex environments. Finally, transitioning from a modular architecture to emerging end-to-end multimodal editing systems may greatly enhance robustness, controllability, and visual fidelity. Overall, this work serves as an initial step toward unified, intelligent visual editing frameworks for creative and advertising applications.

\bibliography{custom}

\clearpage
\appendix

\section*{Appendix}

\section{Prompting Framework and Pipeline Integration}
\label{app:prompting}

This appendix details the prompting strategy used across the full multimodal
object–placement and sponsor–product augmentation pipeline. Each prompt plays a
specific role aligned with the model operating at that stage of the system.

\subsection{LLM Prompting Principles Followed}
\label{app:prompting-principles}

Across all prompts, the following core LLM prompting principles were applied:

\begin{itemize}
    \item \textbf{Clarity and Instruction Explicitness:}  
          Every prompt includes a single, unambiguous task with examples where beneficial.
    \item \textbf{Constrained Output Format:}  
          Prompts strictly specify the expected output format (e.g., ``only the object name'',
          ``comma-separated list'', ``no punctuation''), reducing hallucinations.
    \item \textbf{Context-Rich Grounding:}  
          Prompts explicitly refer to the scene and the region-of-interest (ROI)---in this case,
          the blue bounding box—ensuring image-conditioned reasoning.
    \item \textbf{Negative Guidance:}  
          Diffusion prompts utilise explicit negative prompts to suppress unwanted artefacts
          (e.g., blur, distortion, text).
    \item \textbf{Model-Specific Prompt Alignment:}  
          VLMs receive descriptive reasoning prompts, while generative diffusion models receive
          style-focused prompts.
\end{itemize}

These principles collectively ensure consistency, reliability, and minimal noise throughout the
pipeline.

\section{Prompts by Stage of the Pipeline}
\label{app:prompts}

\subsection{Category Prediction and Object Suggestion}
\label{app:prompts-category-object}

\paragraph{Prompt: Category Prediction}
\begin{quote}
\textit{
Look at the scene in the image. An object should be placed in the blue box. 
From the following list, what are the three most likely categories for an 
object that would fit naturally? Categories: \{CATEGORY\_LIST\}. 
Please respond with only a comma-separated list of the three best category 
names, in order from most likely to least likely.
}
\end{quote}

\paragraph{Prompt: Object Suggestion (LLaVA)}
\begin{quote}
\textit{
The image shows a scene. Considering this scene, suggest a specific object 
for the category '\{category\}' that would fit naturally in the area marked 
by the blue box. Be very specific. For example, if the category is 'Drinks' 
in a cafe, suggest 'a can of Coke' or 'a bottle of water'. Give only the 
object name and nothing else.
}
\end{quote}

\subsection{Branded Object Identification}
\label{app:prompts-branded}

\paragraph{Prompt: Branded Object Finding}
\begin{quote}
\textit{
Analyze this image to find one of the following objects: chips packet, 
soda can, shampoo bottle, perfume can, vacuum cleaner, tshirt, shoes, 
coffee cup. Select the single most visible object from this list. 
Your response must be a short phrase describing the object and its precise 
location in the image followed by a popular brand that fits this object. 
Do not use punctuation.
}
\end{quote}

\subsection{Object Synthesis (SDXL)}
\label{app:prompts-sdxl}

\paragraph{SDXL Generation Prompt}
\begin{quote}
\textit{
cinematic product photo of \{object\_name\}, single object only, clean 
white background, centered, professional studio lighting, immaculate, 
ultra-realistic, 8k, sharp focus
\textbf{Negative prompt:} non white background, blurry, distorted, deformed, 
ugly, low quality, cartoon, text, watermark, signature
}
\end{quote}

\section{Normal Object–Placement Pipeline Illustration}
\label{app:visualization}

To illustrate the multimodal object–placement pipeline, the following figure 
shows the progression through the three intermediate components used in the 
system and how they combine to produce the final augmented output:

\begin{enumerate}
    \item {Original Scene Image}
    \item {Category- and Object-Specific Generated Object}
    \item {Predicted Bounding-Box Localization} 
    \item {Final Composite Output} 
\end{enumerate}

\begin{figure*}[t]
    \centering
    \begin{minipage}{0.30\textwidth}
        \centering
        \includegraphics[width=\linewidth]{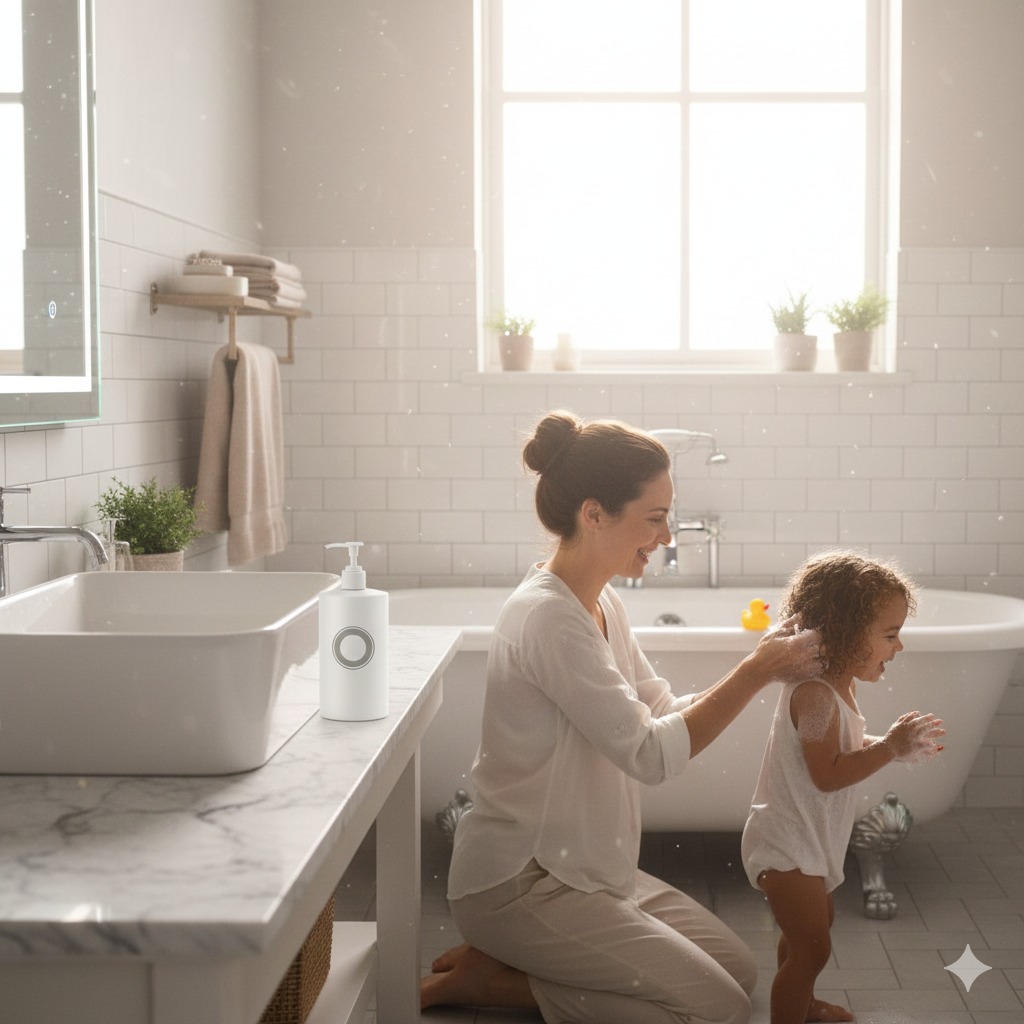}
        \caption*{Source Scene (Video Frame)}
    \end{minipage}
    \hspace{0.5em}
    \large $\mathbf{+}$
    \hspace{0.5em}
    \begin{minipage}{0.20\textwidth}
        \centering
        \includegraphics[width=\linewidth]{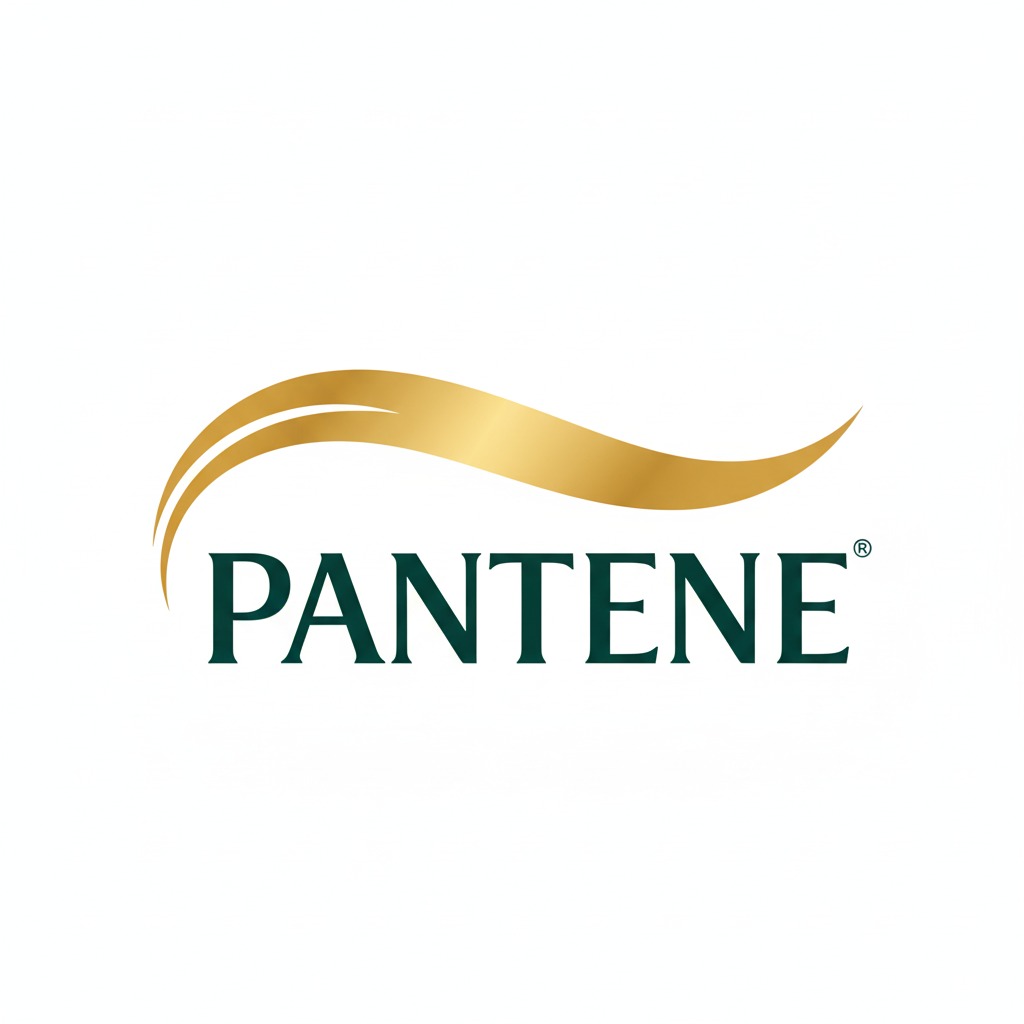}
        \caption*{Sponsor (Logo/Ad)}
    \end{minipage}
    \hspace{0.5em}
    \large $\mathbf{\rightarrow}$
    \hspace{0.5em}
    \begin{minipage}{0.30\textwidth}
        \centering
        \includegraphics[width=\linewidth]{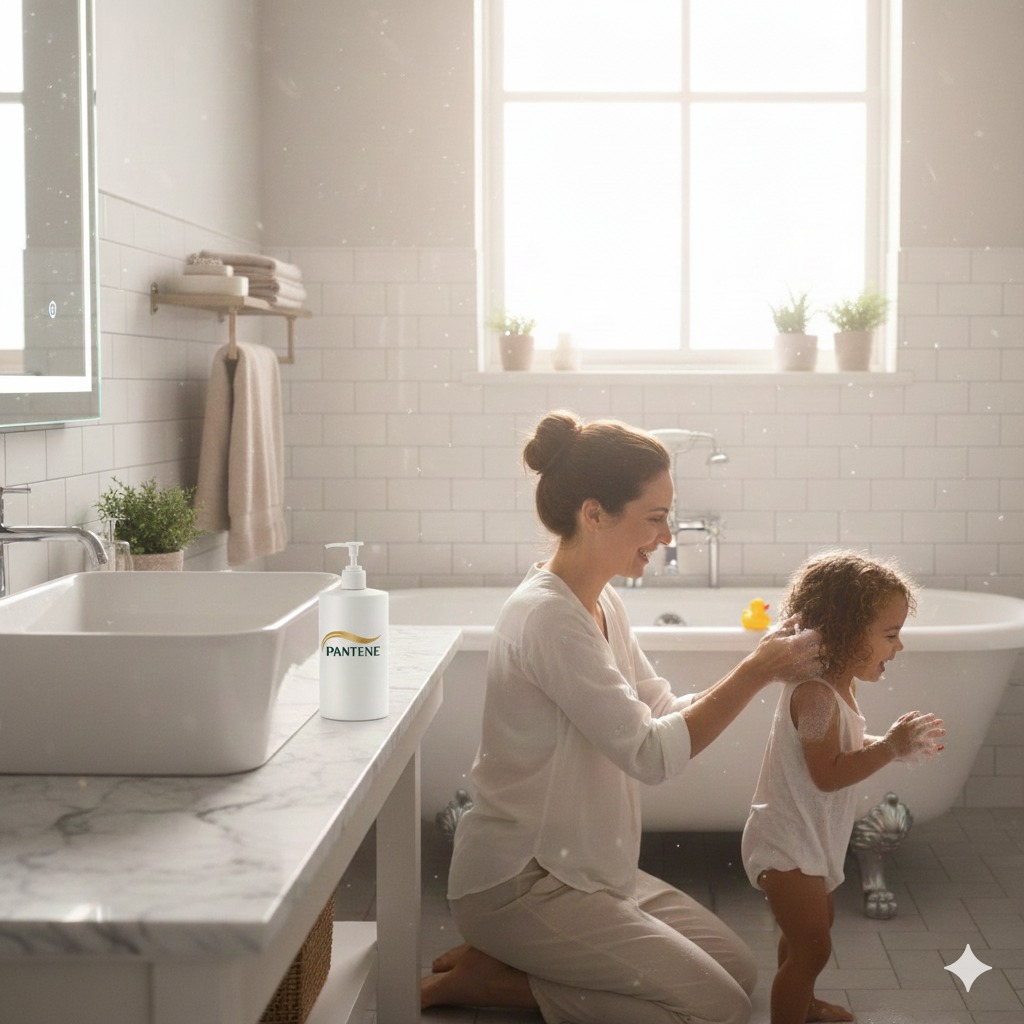}
        \caption*{Final Composite Scene}
    \end{minipage}

    \vspace{0.5em}
    \caption{\textbf{The End-to-End In-Scene Ad Insertion Process.} The figure illustrates the three core elements of the compositing pipeline: the original video frame, the digital sponsor asset (logo), and the final output.}
    \label{fig:in_scene_ad_pipeline}
\end{figure*}

\section{Sponsor-Product Transition Illustration}
\label{app:sponsor}

To demonstrate the sponsor–product augmentation process, the following figure
shows the transition through the three required images:

\begin{enumerate}
    \item Original bathroom scene  
    \item Branded PANTENE product logo  
    \item Final augmented scene with the brand-infused object  
\end{enumerate}
\end{document}